\documentclass{article}

\usepackage{microtype}
\usepackage{graphicx}

\usepackage{hyperref}

\usepackage[accepted]{icml2026}
\usepackage{booktabs}       
\usepackage{amsfonts}
\usepackage[table]{xcolor}
\usepackage{nicefrac}       
\usepackage{microtype}      
\definecolor{boxcolor}{HTML}{B85450}
\newcommand{\red}[1]{{\color{red}#1}}

\newcommand{\cross}{\textcolor{red}{\ding{55}}}

\usepackage{graphicx}
\usepackage{booktabs}
\usepackage{multirow}
\usepackage{enumitem}
\usepackage{comment}
\usepackage{caption} 
\usepackage{arydshln}
\usepackage{tikz}       
\usepackage{adjustbox}  
\usepackage{circledsteps}
\usepackage{amsmath}
\usepackage{wrapfig}
\usepackage{colortbl}
\usepackage{tabularray}
\usepackage{caption}
\usepackage{subcaption}
\usepackage{nicematrix}
\usepackage{amssymb}
\usepackage{bbm}
\usepackage{lipsum}
\usepackage{svg}
\usepackage{wrapfig}
\usepackage{pifont}

\usepackage{hyperref}
\usepackage{url}
\hypersetup{
colorlinks=true,
}

\usepackage[capitalize,noabbrev]{cleveref}

\usepackage[disable,textsize=tiny]{todonotes}

\icmltitlerunning{Listen, Look, and Learn: Learning Without Forgetting through SAM-Audio}

\begin{document}

\twocolumn[
  \icmltitle{Listen, Look, and Learn: Learning Without Forgetting through SAM-Audio}



  \icmlsetsymbol{equal}{*}

  \begin{icmlauthorlist}
    \icmlauthor{Avi Gupta}{yyy}
    \icmlauthor{Nilotpal Sinha}{comp}
    \icmlauthor{Vishnu Raj}{comp}
    \icmlauthor{Sambuddha Saha}{comp}
    \icmlauthor{Pratik Joshi}{comp}
    \icmlauthor{Koteswar Rao Jerripothula}{sch,yyy}
    \icmlauthor{Tammam Tillo}{sch2,yyy}
  \end{icmlauthorlist}

  \icmlaffiliation{yyy}{IIIT Delhi}
  \icmlaffiliation{comp}{Dolby Laboratories}
  \icmlaffiliation{sch}{IIT Kanpur}
  \icmlaffiliation{sch2}{TYUST China}

  \icmlcorrespondingauthor{Avi Gupta}{avig@iiitd.ac.in}


  \vskip 0.3in
]



\printAffiliationsAndNotice{This work was done by Avi Gupta during his internship at Dolby Laboratories.}  

\begin{abstract}
Class-Incremental Learning (CIL) aims to continuously learn new classes without forgetting previously acquired knowledge. While recent CIL advances have spurred significant interest across various modalities, the audio-visual setting remains underexplored. Furthermore, although foundational multimodal models like SAM-Audio encapsulate rich static priors, our empirical analysis reveals that these representations struggle in incremental settings. This work bridges this gap by integrating SAM-Audio's audio-visual priors into the CIL setting. Specifically, we leverage its dense audio and visual representations and employ a novel guided attention strategy where the audio features contextually guide the visual representations. To further mitigate catastrophic forgetting, we introduce dual-level distillation objectives at both the feature and logit levels. Extensive evaluations on audio-visual CIL benchmarks demonstrate that our approach consistently outperforms state-of-the-art methods.
\end{abstract}
\begin{figure}[htb]
\vspace{-10pt}
    \centering
    \includegraphics[width=\columnwidth]{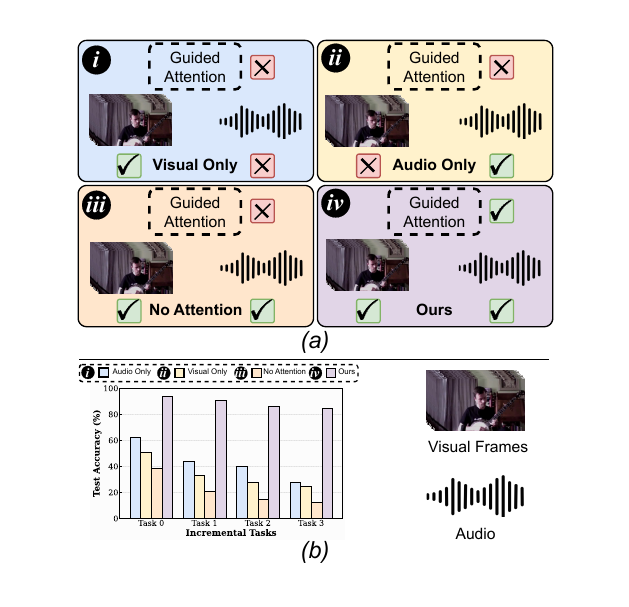}
    \vspace{-20pt}
    \caption{\footnotesize Overview of our proposed setting on the AVE dataset. (a) We present the four approaches: {\small \Circled[fill color=black, inner color=white]{\textit{\textbf{i}}}} only visual modality; {\small \Circled[fill color=black, inner color=white]{\textit{\textbf{ii}}}} only audio modality; {\small \Circled[fill color=black, inner color=white]{\textit{\textbf{iii}}}} both modalities fused naively (baseline); {\small \Circled[fill color=black, inner color=white]{\textit{\textbf{iv}}}} our proposed approach with guided attention-based fusion. (b) The superior performance of our method over individual modalities and the baseline underscores the necessity of coupling multimodal representations with guided attention for robust class-incremental learning.
     }
    \label{fig:teaser}
    \vspace{-10pt}
\end{figure}
\section{Introduction}
Class-Incremental Learning (CIL) aims to learn new classes incrementally without forgetting the previously learned ones. Prior works have leveraged different strategies including knowledge distillation~\cite{castro2018end, dhar2019learning, douillard2020podnet}, adapter-based~\cite{He_2025_CVPR, wang2025integrating}, data/memory-replay~\cite{qi2025class, channappayya2023augmented, chen2023saving}, etc. The CIL setting is widely used across diverse practical applications, including robotics~\cite{yao2025think}, autonomous driving, and medical imaging~\cite{yi2025idpa}, where learnable systems must adapt to new objects or diseases without retraining from scratch. Despite significant advances, understanding class-incremental learning in foundational multimodal learning remains limited.

Foundational models~\cite{radford2021learning, achiam2023gpt, abdin2024phi, kirillov2023segment} are large-scale models pre-trained on vast, diverse datasets to act as a general-purpose base for various downstream applications. They have shown great advancements in low-level vision tasks~\cite{yadav2025samwave}, virtual assistants~\cite{kumar2025gpt}, autonomous driving~\cite{guan2024world}, medical imaging~\cite{azad2023foundational}, etc.
Multimodal learning focuses on targeting two or more different modalities, from language, vision, and audio, to interact and provide a better understanding of the scene. Recent works in vision-language continual models~\cite{jha2024clap4clip, cao2024continual} have focused on recognizing new classes while retaining strong, pre-trained capabilities. Similarly, in audio-visual modalities, some methods \cite{pian2023audio, DBLP:conf/nips/PianNDMGT24, DBLP:journals/corr/abs-2510-17234} target to fuse both audio and visual feature representations to understand the scene in an incremental setting. However, these representations are extracted from separate encoders and require an additional strategy to align them in a common space. To address that, we propose leveraging SAM-Audio~\cite{shi2025samaudio}, a foundational multimodal model that already has this innate ability due to its pre-training.

To assess the robustness of SAM-Audio's multimodal representational space in the incremental setting, we conduct a preliminary empirical analysis of its audio and visual features. 
We compare four distinct configurations, as illustrated in Fig.~\ref{fig:teaser}(a): {\small \Circled[fill color=black, inner color=white]{\textit{\textbf{i}}}} unimodal visual features; {\small \Circled[fill color=black, inner color=white]{\textit{\textbf{ii}}}} unimodal audio features; {\small \Circled[fill color=black, inner color=white]{\textit{\textbf{iii}}}} a baseline utilizing naive audio-visual fusion; {\small \Circled[fill color=black, inner color=white]{\textit{\textbf{iv}}}} our proposed guided attention-based fusion; The empirical results in Fig.~\ref{fig:teaser}(b) reveal that relying on isolated unimodal representations yields severe performance degradation across successive incremental tasks. Crucially, the naive fusion baseline fails to mitigate catastrophic forgetting, resulting in suboptimal overall accuracy. These findings demonstrate that simply leveraging individual modalities or employing naive fusion strategies are insufficient for CIL.

To address these limitations, we introduce a guided strategy designed to leverage SAM-Audio's frozen representations for CIL. While large multimodal foundation models encapsulate rich audio-visual priors, our approach effectively leverages these static latent spaces to sequential learning tasks. Specifically, we extract features from the frozen SAM-Audio's audio and visual encoders and process them through a guided attention mechanism. In this module, audio representations serve as a contextual signal to guide the visual features toward precise target predictions. To further mitigate catastrophic forgetting, we employ strict knowledge distillation objectives at both the feature and logit levels. This ensures that the topology of previously learned multimodal alignments is preserved as new classes are introduced. As shown in Fig.~\ref{fig:teaser}(b), our proposed approach consistently outperforms the baselines across all incremental tasks. We summarize our key contributions as follows: 
{\small \Circled[fill color=black, inner color=white]{\textit{\textbf{1}}}} We propose a novel, efficient paradigm for adapting large-scale, pre-trained audio-visual foundation models (specifically SAM-Audio) to the challenging class-incremental learning setting, bypassing the need for computationally expensive full fine-tuning. {\small \Circled[fill color=black, inner color=white]{\textit{\textbf{2}}}} We introduce a specialized guided attention mechanism that leverages dense audio representations as a conditioning signal to dynamically adapt visual features for sequential semantic prediction. {\small \Circled[fill color=black, inner color=white]{\textit{\textbf{3}}}} To explicitly mitigate catastrophic forgetting, we formulate a robust distillation objective operating at both the feature and logit levels, ensuring the preservation of previously learned classes as new classes are introduced. {\small \Circled[fill color=black, inner color=white]{\textit{\textbf{4}}}} Extensive experiments demonstrate that our proposed framework significantly outperforms existing state-of-the-art methods across standard benchmarks.
\begin{figure*}[h] 
\centering
\includegraphics[width=.8\textwidth]{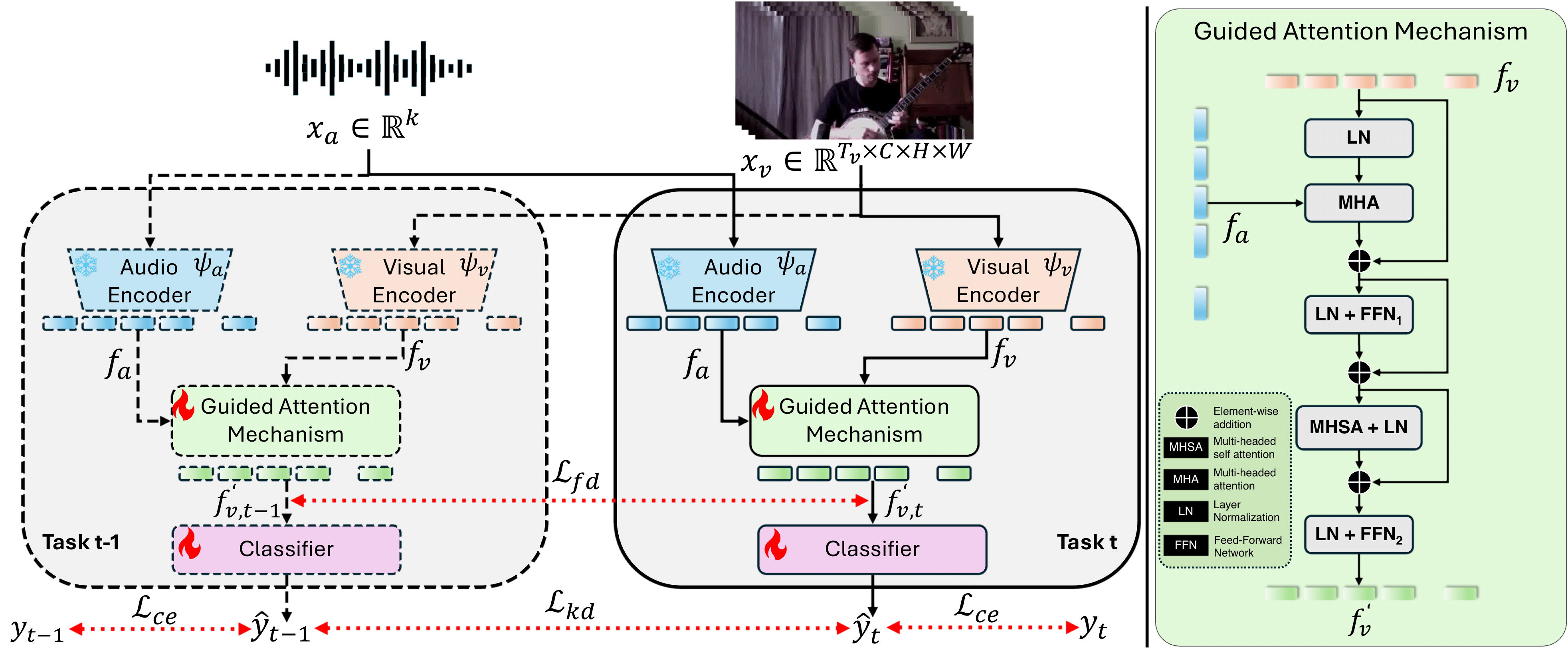}
\vspace{-5pt}
\caption{\footnotesize \textbf{[Left]} The overall architecture. At current task $t$, audio ($f_a$) and visual ($f_v$) representations are derived from audio signal ($x_a$) and video frames ($x_v$) using frozen audio ($\psi_a$) and visual ($\psi_v$) encoders, respectively of the SAM-Audio. These audio and frame embeddings are fused by a guided attention mechanism. The final video representations ($f'_v$) are passed through the classifier for precise class prediction. The model is trained using $\mathcal{L}_{ce}$ (for class prediction), $\mathcal{L}_{fd}$ (distillation at feature level), and $\mathcal{L}_{kd}$ (distillation at logits level). 
\textbf{[Right]} The guided attention mechanism, comprising a combination of multi-headed cross attention and multi-headed self attention, takes the audio representations and provides spatial and semantic guidance to the visual representation for precise class prediction.}
\label{fig:arch}
\vspace{-10pt}
\end{figure*}

\section{Proposed Method}
\subsection{Overview}
The objective is to leverage the dense representations of the foundational multimodal model, SAM-Audio, and strategically adapt to new, unknown classes in an incremental setting while avoiding forgetting previously learned ones. The incremental setting is structured as a sequence of tasks $t\in \{0,1,..., T\}$. Each task $t$ provides a distinct set of classes $C_t$. Following~\cite{pian2023audio}, the training dataset for the task $t$ includes $\mathcal{D}'_t = \mathcal{D}_t\cup \mathcal{M}_{t-1}$, where $\mathcal{D}_t$ includes all the samples from current task while $\mathcal{M}_{t-1}$ comprises exemplar samples from the previous task. These exemplar samples are randomly selected from all tasks up to $t-1$. Each sample consists of a video sample $x$ and its corresponding label $y$.

Our proposed approach is divided into three major components: \textit{(a) Multimodal Feature Extractor}, that extracts the audio-visual features from multimodal encoder~\cite{shi2025samaudio}; \textit{(b) Guided Attention Mechanism}, that fuses the extracted audio and visual features using transformer-based attention and provides sample-specific multimodal composite features; and \textit{(c) Classifier}, that finally predicts the logits from the multimodal features. The overview of our proposed architecture is visualized in Fig.~\ref{fig:arch}.

\subsection{SAM-Audio as the Backbone}

As mentioned before, the objective of our proposed approach is to leverage the learned representations of SAM-Audio for the incremental setting. Hence, we use the SAM-Audio as the backbone for both audio and visual modality. The video encoder of the SAM-Audio is
trained 
on over 100 million videos. 
We also leverage the SAM-Audio's audio encoder. The audio encoder is based on the DAC-like autoencoder structure with a Variational Autoencoder (VAE) bottleneck layer. The audio is encoded into a sequence at 25 Hz. As in the video encoder, we keep the weights of the pretrained audio encoder frozen as well.

Given an input video sample $x$, we extract the video frames and audio signal $(x = \{x_v, x_a\})$ where, $x_v \in \mathbb{R}^{T_v\times C\times H\times W}$ and audio signal $x_a \in \mathbb{R}^{k}$. We use the visual $\psi_v$ and audio $\psi_a$ encoder to generate the corresponding visual features $f_v \in \mathbb{R}^{T_v\times d}$ and audio features $f_a \in \mathbb{R}^{T_a\times d}$ as: $f_v = \psi_v(x_v);$ and $f_a = \psi_a(x_a)$, respectively.

Here, $T_a$ and $T_v$ are the audio chunks and visual frames, respectively. $C$, $H$, and $W$ are the number of channels, the height, and the width of each frame, respectively. $d$ represents the hidden dimension. These extracted features are passed through the transformer-based attention mechanism to leverage the audio representation for video classification.

\subsection{Guided Attention Mechanism}
The audio and visual embeddings from multimodal encoders provide robust and aligned information about the corresponding inputs. To effectively integrate these embeddings, inspired by ~\cite{Jeong_2025_CVPR}, we leverage a guided attention mechanism. This transformer-based attention provides effective guidance from audio, thus enhancing the visual content. The guided attention, visualized in Fig.~\ref{fig:arch}[Right], consists of an attention block with a Multi-Head Attention (MHA) and a Feed-Forward Network (FFN) to blend the audio and frame representations. This is followed by Multi-Head Self-Attention (MHSA) and a Feed-Forward Network (FFN) for the representation refinement. 

In particular, the attention block takes the audio $f_a$ and frame $f_v$ representations as input and fuses them together by MHA. This process is formulated as:  
\begin{equation}
    \begin{aligned}
        z &= \text{MHA}(\text{LN}(f_v), \text{LN}(f_a)) + f_v; \\
        \overline{z} &= \text{FFN}_1(\text{LN}(z)) + z\\
    \end{aligned}
\end{equation}
where LN denotes the layer normalization. The output representation $\overline{z}$ is refined to enhance overall visual quality. Specifically, $\overline{z}$ is passed through the Multi-Head Self-Attention (MHSA)
module followed by another FFN to produce the refined visual representation 
$f'_v$. The refining process is formulated as:
\begin{equation}
    \begin{aligned}
        \tilde{z} &= \text{MHSA}(\text{LN}(\overline{z})) + \overline{z};\\
        f'_v &= \text{FFN}_2(\text{LN}(\tilde{z})) + \tilde{z}
    \end{aligned}
\end{equation}
\subsection{Objective Function}
Our proposed strategy employs \textit{task-separated cross-entropy} $\mathcal{L}_{ce}$ for the class-incremental learning, which also prevents the prediction of old classes from being updated by learning new classes. The loss is formulated as:
\begin{equation}
\mathcal{L}_{ce} = \sum_{i=1}^{\mathcal{C}_t}y_i\log(\hat{y}_i)
\end{equation}
where $\hat{y}_i$, and $y_i$ denotes the predicted and ground-truth labels respectively. Also, we use the \textit{feature distillation loss} $\mathcal{L}_{fd}$ between the attended visual features of the current and the previous task (on the exemplar samples $\mathcal{M}_{t-1}$) to teach $f'_{v,t}$ to mimic $f'_{v,t-1}$. It is formulated as:
\begin{equation}
\mathcal{L}_{fd} = ||f'_{v,t}-f'_{v,t-1}||^2
\end{equation}
The last part of our final objective function is the \textit{task-wise knowledge distillation} $\mathcal{L}_{kd}$, where we evaluate the Kullback-Leibler (KL) divergence between the logits of classes in the current ($\hat{y}_t^{C_s}$) and previous ($\hat{y}_{t-1}^{C_s}$) tasks which helps preserve past knowledge and prevent the learned knowledge from being biased by other tasks.
\begin{equation}
\mathcal{L}_{kd} = \sum_{s=1}^t KL(\hat{y}_t^{C_s} || \hat{y}_{t-1}^{C_s})
\end{equation}
To sum it up, our overall \textbf{\textit{objective function}} is formulated as: $\mathcal{L}_{tot} = \mathcal{L}_{ce} +\mathcal{L}_{fd} + \mathcal{L}_{kd}$

\section{Experiments}
\subsection{Experimental Details}
\textbf{Datasets: }For fair comparison, following~\cite{pian2023audio}, we evaluate our proposed approach on AVE-Class-Incremental (AVE-CI) dataset which comprises 4K+ videos covering 28 audio-visual event classes that are uniformly distributed into 4 incremental tasks, with 7 classes each; and VGGSound100-Class-Incremental (VS100-CI) dataset with 100 classes and 60K samples that are distributed into 10 incremental tasks, with 10 classes in each.\\
\textbf{Evaluation Metrics: }We evaluate the performance of our approach with Mean Accuracy:
\begin{equation}
    MeanAcc = \frac{1}{T}\sum_{t=1}^{T}a_t
\end{equation}
where $a_t$ is the testing accuracy of all seen classes after completing the training on the current task $t$.

We also evaluate the average forgetting to measure the extent of catastrophic forgetting over previously learned tasks:
\begin{equation}
\begin{aligned}
    AvgForgetting &= \frac{1}{T-1}\sum_{t=2}^{T}F_t \\
    F_t = \frac{1}{t-1}\sum_{i=1}^{t-1}& \max_{k\in\{i,...,t-1\}}(a_{k,i}- a_{t,i})
\end{aligned}
\end{equation}
where $a_{k,i}$ is the testing accuracy of the i-th task after training on the k-th task.\\
\textbf{Implementation Details: }We leverage the foundational SAM-Audio for the audio and visual encoder. We freeze the pre-trained visual and audio encoders during training and fine-tune the rest of the model (audio-guided attention and the classifier). We use the SGD optimizer with a learning rate of 5e-5 and a weight decay of 1e-4, and use Cosine Annealing for the learning rate decay. For the AVE-CI and VS100-CI datasets, we set the maximum number of training epochs per incremental step to 200.
\begin{table}[h]
\vspace{-5pt}
\centering
\caption{\footnotesize Comparison of our proposed approach with current state-of-the-art methods. \textbf{Bold} denotes the best results.}
\vspace{-5pt}
\label{tab:results}
\renewcommand{\arraystretch}{1}
\resizebox{\columnwidth}{!}{
\setlength{\tabcolsep}{3pt}
\begin{tabular}{lcccc}
\toprule
\midrule
\multirow{2}{*}{Methods} & \multicolumn{2}{c}{AVE-CI} & \multicolumn{2}{c}{VS100-CI} \\
\cmidrule(lr){2-3} \cmidrule(lr){4-5}
& Mean Acc.$\uparrow$ & Avg. Forgetting$\downarrow$ & Mean Acc.$\uparrow$ & Avg. Forgetting$\downarrow$ \\
\midrule
Fine-tuning           & 42.40 & 70.99 & 26.21 & 89.37 \\
LwF~\cite{8107520}                   & 58.07 & 26.90 & 59.34 & 23.01 \\
iCaRL-NME~\cite{8100070}             & 56.15 & 11.71 & 56.19 & 12.80 \\
iCaRL-FC~\cite{8100070}              & 65.88 & 26.08 & 64.22 & 29.94 \\
SS-IL~\cite{9710553}                 & 61.94 & 22.49 & 69.20 & 9.75  \\
AFC-NME~\cite{Kang2022afc}               & 68.46 & 14.18 & 61.41 & 23.30 \\
AFC-LSC~\cite{Kang2022afc}               & 65.21 & 28.11 & 57.76 & 29.64 \\
AV-CIL~\cite{pian2023audio}                & 74.04 & 7.63  & 72.80 & 5.49  \\ \midrule
Ours                  & \textbf{88.72} & \textbf{2.14}  & \textbf{79.64} & \textbf{0.34} \\
\midrule
\bottomrule
\end{tabular}
}
\vspace{-15pt}
\end{table}

\subsection{Results and Analysis: }
In this section, we compare our proposed approach for audio-visual CIL with the current state of the art. The results are illustrated in Table~\ref{tab:results}. From the table, we observe that our proposed approach surpasses the previous works by a significant margin. 
Particularly, on the AVE-CI dataset, our approach consistently outperforms previous state-of-the-art ~\cite{pian2023audio} in both mean test accuracy and average forgetting. Similar performance gains are observed on VS100-CI, highlighting the robustness of our method across benchmarks.\\

\subsection{Ablation Studies}
\textbf{Effectiveness of loss functions. }In Table~\ref{tab:ablation_1}, we demonstrate the effectiveness of our individual loss functions on the AVE-CI dataset. We observe that removing both the feature ($\mathcal{L}_{fd}$) and knowledge distillation ($\mathcal{L}_{kd}$) losses results in poor test accuracy and further increases the risk of forgetting. To identify the individual contribution of both distillation-based losses, we remove $\mathcal{L}_{fd}$ and $\mathcal{L}_{kd}$ separately. We observe that including only $\mathcal{L}_{kd}$ improves performance, and including $\mathcal{L}_{fd}$ demonstrates \textit{negative forgetting}, indicating that the model does not actually forget prior knowledge. Hence, both losses play an important role in enhancing overall performance and mitigating catastrophic forgetting.

\begin{table}[htbp] 
\centering

\begin{minipage}{0.485\columnwidth}
\centering
\caption{\footnotesize Ablation study to analyze the effectiveness of loss components and guided attention. \textbf{Bold} denotes the best results.}
\vspace{-5pt}
\label{tab:ablation_1}
\renewcommand{\arraystretch}{1}
\resizebox{\linewidth}{!}{
\setlength{\tabcolsep}{5.5pt}
\begin{tabular}{cccccc}
\toprule
\midrule
Guided & $\mathcal{L}_{ce}$ & $\mathcal{L}_{kd}$ & $\mathcal{L}_{fd}$  & Mean & Avg. \\
Attention & &  & & Acc.$\uparrow$ & Forgetting$\downarrow$\\
\midrule
\cross  & \checkmark & \checkmark      &   \cross    & 21.45 & 12.58 \\ 
\checkmark  & \checkmark & \cross  & \cross   & 87.26  & 4.30  \\
\checkmark  & \checkmark & \checkmark & \cross   & 87.34 & 2.29  \\
\checkmark  & \checkmark & \cross     & \checkmark    & 87.79 & \textbf{-1.32} \\ \midrule
\checkmark  &   \checkmark &   \checkmark &  \checkmark    & \textbf{88.72} & 2.14 \\
\midrule
\bottomrule
\end{tabular}
}
\end{minipage}\hfill 
\begin{minipage}{0.485\columnwidth}
\centering
\caption{\footnotesize Ablation study to analyze the effectiveness of individual modality on AVE dataset. \textbf{Bold} denotes the best results.}
\vspace{-5pt}
\label{tab:ablation_2}
\renewcommand{\arraystretch}{1}
\resizebox{\linewidth}{!}{
\setlength{\tabcolsep}{1.5pt}
\begin{tabular}{cccc}
\toprule
\midrule
\multicolumn{2}{c}{Modality} & \multirow{2}{*}{Mean Acc.$\uparrow$} & \multirow{2}{*}{Avg. Forgetting$\downarrow$} \\
\cmidrule(lr){1-2}
Audio & Video & & \\
\midrule
\checkmark & \cross & 34.21 & 11.30  \\ 
\cross & \checkmark & 43.50	& 5.66  \\ \midrule
\checkmark & \checkmark & \textbf{88.72} & \textbf{2.14}  \\
\midrule
\bottomrule
\end{tabular}%
}
\end{minipage}
\vspace{-10pt}
\end{table}
\textbf{Effectiveness of individual modality. }In Table~\ref{tab:ablation_2}, we perform another study to analyze the effectiveness of both audio and visual modalities in the class-incremental setting. We observe that using both modalities significantly improves performance and reduces forgetting compared to using either modality alone. Furthermore, we see some improvement when using only the visual modality over audio, both in accuracy and in forgetting; however, the individual modalities do not keep up with our approach. The per-step test accuracy of individual audio and visual modality is also visualized in Fig.~\ref{fig:teaser}(b). Additional analysis and ablation studies are provided in the \red{\textbf{\textit{Appendix}}}.
\section{Conclusion}
In this work, we explore the task of continual learning using foundational multimodal models to effectively learn new visual classes with audio guidance, without catastrophically forgetting previously learned knowledge. To address the proposed problem, we leverage the dense representations of the foundational multimodal model, SAM-Audio. Furthermore, we propose an audio-guided transformer-based attention mechanism. Finally, we utilize knowledge distillation at the task and feature levels to retain prior knowledge. Experiments on benchmark datasets show that our approach outperforms the current state-of-the-art methods by a significant margin. Given the significant performance improvement from SAM-Audio, we plan to generalize our proposed approach to other foundational multimodal models as future work.
\clearpage
\bibliography{mybib}

@inproceedings{pian2023audio,
  title={Audio-visual class-incremental learning},
  author={Pian, Weiguo and Mo, Shentong and Guo, Yunhui and Tian, Yapeng},
  booktitle={Proceedings of the IEEE/CVF International Conference on Computer Vision},
  pages={7799--7811},
  year={2023}
}

@inproceedings{DBLP:conf/nips/PianNDMGT24,
  author       = {Weiguo Pian and
                  Yiyang Nan and
                  Shijian Deng and
                  Shentong Mo and
                  Yunhui Guo and
                  Yapeng Tian},
  editor       = {Amir Globersons and
                  Lester Mackey and
                  Danielle Belgrave and
                  Angela Fan and
                  Ulrich Paquet and
                  Jakub M. Tomczak and
                  Cheng Zhang},
  title        = {Continual Audio-Visual Sound Separation},
  booktitle    = {Advances in Neural Information Processing Systems 38: Annual Conference
                  on Neural Information Processing Systems 2024, NeurIPS 2024, Vancouver,
                  BC, Canada, December 10 - 15, 2024},
  year         = {2024},
  bibsource    = {dblp computer science bibliography, https://dblp.org}
}

@article{DBLP:journals/corr/abs-2510-17234,
  author       = {Yuyang Hong and
                  Qi Yang and
                  Tao Zhang and
                  Zili Wang and
                  Zhaojin Fu and
                  Kun Ding and
                  Bin Fan and
                  Shiming Xiang},
  title        = {Taming Modality Entanglement in Continual Audio-Visual Segmentation},
  journal      = {CoRR},
  volume       = {abs/2510.17234},
  year         = {2025},
  url          = {https://doi.org/10.48550/arXiv.2510.17234},
  doi          = {10.48550/ARXIV.2510.17234},
  eprinttype    = {arXiv},
  eprint       = {2510.17234},
  bibsource    = {dblp computer science bibliography, https://dblp.org}
}

@article{shi2025samaudio,
    title={SAM Audio: Segment Anything in Audio},
    author={Bowen Shi and Andros Tjandra and John Hoffman and Helin Wang and Yi-Chiao Wu and Luya Gao and Julius Richter and Matt Le and Apoorv Vyas and Sanyuan Chen and Christoph Feichtenhofer and Piotr Doll{\'a}r and Wei-Ning Hsu and Ann Lee},
    year={2025},
    url={https://arxiv.org/abs/2512.18099}
}

@InProceedings{Jeong_2025_CVPR,
    author    = {Jeong, Boseung and Park, Jicheol and Kim, Sungyeon and Kwak, Suha},
    title     = {Learning Audio-guided Video Representation with Gated Attention for Video-Text Retrieval},
    booktitle = {Proceedings of the IEEE/CVF Conference on Computer Vision and Pattern Recognition (CVPR)},
    month     = {June},
    year      = {2025},
    pages     = {26202-26211}
}

@article{He_2025_CVPR,
    author    = {He, Jiangpeng and Duan, Zhihao and Zhu, Fengqing},
    title     = {CL-LoRA: Continual Low-Rank Adaptation for Rehearsal-Free Class-Incremental Learning},
    journal = {Proceedings of the Computer Vision and Pattern Recognition Conference (CVPR)},
    month     = {June},
    year      = {2025},
    pages     = {30534-30544}
}

@inproceedings{wang2025integrating,
  title={Integrating Task-Specific and Universal Adapters for Pre-Trained Model-based Class-Incremental Learning},
  author={Yan Wang and Da-Wei Zhou and Han-Jia Ye},
  booktitle={ICCV},
  year={2025}
}

@article{channappayya2023augmented,
  title={Augmented memory replay-based continual learning approaches for network intrusion detection},
  author={Channappayya, Sumohana and Tamma, Bheemarjuna Reddy and others},
  journal={Advances in Neural Information Processing Systems},
  volume={36},
  pages={17156--17169},
  year={2023}
}

@article{chen2023saving,
  title={Saving 100x storage: Prototype replay for reconstructing training sample distribution in class-incremental semantic segmentation},
  author={Chen, Jinpeng and Cong, Runmin and Luo, Yuxuan and Ip, Horace and Kwong, Sam},
  journal={Advances in Neural Information Processing Systems},
  volume={36},
  pages={35988--35999},
  year={2023}
}

@article{qi2025class,
  title={Class-wise Balancing Data Replay for Federated Class-Incremental Learning},
  author={Qi, Zhuang and Tang, Ying-Peng and Meng, Lei and Yu, Han and Li, Xiaoxiao and Meng, Xiangxu},
  journal={arXiv preprint arXiv:2507.07712},
  year={2025}
}

@inproceedings{yao2025think,
  title={Think Small, Act Big: Primitive Prompt Learning for Lifelong Robot Manipulation},
  author={Yao, Yuanqi and Liu, Siao and Song, Haoming and Qu, Delin and Chen, Qizhi and Ding, Yan and Zhao, Bin and Wang, Zhigang and Li, Xuelong and Wang, Dong},
  booktitle={Proceedings of the Computer Vision and Pattern Recognition Conference},
  pages={22573--22583},
  year={2025}
}

@article{yi2025idpa,
  title={iDPA: Instance Decoupled Prompt Attention for Incremental Medical Object Detection},
  author={Yi, Huahui and Xu, Wei and Qin, Ziyuan and Chen, Xi and Wu, Xiaohu and Li, Kang and Lao, Qicheng},
  journal={arXiv preprint arXiv:2506.00406},
  year={2025}
}

@inproceedings{Yadav_2025_BMVC,
author    = {Saurabh Yadav and Avi Gupta and Koteswar Rao Jerripothula},
title     = {SAMWave: Adapting Segment Anything Model to difficult tasks},
booktitle = {36th British Machine Vision Conference 2025, {BMVC} 2025, Sheffield, UK, November 24-27, 2025},
publisher = {BMVA},
year      = {2025},
url       = {https://bmva-archive.org.uk/bmvc/2025/assets/papers/Paper_698/paper.pdf}
}

@article{yadav2025samwave,
  title={SAMwave: Wavelet-Driven Feature Enrichment for Effective Adaptation of Segment Anything Model},
  author={Yadav, Saurabh and Gupta, Avi and Jerripothula, Koteswar Rao},
  journal={arXiv preprint arXiv:2507.20186},
  year={2025}
}

@ARTICLE{8107520,
  author={Li, Zhizhong and Hoiem, Derek},
  journal={IEEE Transactions on Pattern Analysis and Machine Intelligence}, 
  title={Learning without Forgetting}, 
  year={2018},
  volume={40},
  number={12},
  pages={2935-2947},
  doi={10.1109/TPAMI.2017.2773081}}

@INPROCEEDINGS{8100070,
  author={Rebuffi, Sylvestre-Alvise and Kolesnikov, Alexander and Sperl, Georg and Lampert, Christoph H.},
  booktitle={2017 IEEE Conference on Computer Vision and Pattern Recognition (CVPR)}, 
  title={iCaRL: Incremental Classifier and Representation Learning}, 
  year={2017},
  volume={},
  number={},
  pages={5533-5542},
  doi={10.1109/CVPR.2017.587}}

@INPROCEEDINGS{9710553,
  author={Ahn, Hongjoon and Kwak, Jihwan and Lim, Subin and Bang, Hyeonsu and Kim, Hyojun and Moon, Taesup},
  booktitle={2021 IEEE/CVF International Conference on Computer Vision (ICCV)}, 
  title={SS-IL: Separated Softmax for Incremental Learning}, 
  year={2021},
  volume={},
  number={},
  pages={824-833},
  doi={10.1109/ICCV48922.2021.00088}}

@inproceedings{Kang2022afc,
  author = {Kang, Minsoo and Park, Jaeyoo and Han, Bohyung},
  booktitle = {CVPR},
  title = "{Class-Incremental Learning by Knowledge Distillation with Adaptive Feature Consolidation}",
  year = {2022}
  }

@article{ahn2019uncertainty,
  title={Uncertainty-based continual learning with adaptive regularization},
  author={Ahn, Hongjoon and Cha, Sungmin and Lee, Donggyu and Moon, Taesup},
  journal={Advances in neural information processing systems},
  volume={32},
  year={2019}
}

@article{benjamin2018measuring,
  title={Measuring and regularizing networks in function space},
  author={Benjamin, Ari S and Rolnick, David and Kording, Konrad},
  journal={arXiv preprint arXiv:1805.08289},
  year={2018}
}

@InProceedings{Chaudhry_2018_ECCV,
author = {Chaudhry, Arslan and Dokania, Puneet K. and Ajanthan, Thalaiyasingam and Torr, Philip H. S.},
title = {Riemannian Walk for Incremental Learning: Understanding Forgetting and Intransigence},
booktitle = {Proceedings of the European Conference on Computer Vision (ECCV)},
month = {September},
year = {2018}
}

@article{jung2020continual,
  title={Continual learning with node-importance based adaptive group sparse regularization},
  author={Jung, Sangwon and Ahn, Hongjoon and Cha, Sungmin and Moon, Taesup},
  journal={Advances in neural information processing systems},
  volume={33},
  pages={3647--3658},
  year={2020}
}

@inproceedings{castro2018end,
  title={End-to-end incremental learning},
  author={Castro, Francisco M and Mar{\'\i}n-Jim{\'e}nez, Manuel J and Guil, Nicol{\'a}s and Schmid, Cordelia and Alahari, Karteek},
  booktitle={Proceedings of the European conference on computer vision (ECCV)},
  pages={233--248},
  year={2018}
}

@inproceedings{dhar2019learning,
  title={Learning without memorizing},
  author={Dhar, Prithviraj and Singh, Rajat Vikram and Peng, Kuan-Chuan and Wu, Ziyan and Chellappa, Rama},
  booktitle={Proceedings of the IEEE/CVF conference on computer vision and pattern recognition},
  pages={5138--5146},
  year={2019}
}

@inproceedings{douillard2020podnet,
  title={Podnet: Pooled outputs distillation for small-tasks incremental learning},
  author={Douillard, Arthur and Cord, Matthieu and Ollion, Charles and Robert, Thomas and Valle, Eduardo},
  booktitle={European conference on computer vision},
  pages={86--102},
  year={2020},
  organization={Springer}
}

@inproceedings{hou2019learning,
  title={Learning a unified classifier incrementally via rebalancing},
  author={Hou, Saihui and Pan, Xinyu and Loy, Chen Change and Wang, Zilei and Lin, Dahua},
  booktitle={Proceedings of the IEEE/CVF conference on computer vision and pattern recognition},
  pages={831--839},
  year={2019}
}

@article{chaudhry2018efficient,
  title={Efficient lifelong learning with a-gem},
  author={Chaudhry, Arslan and Ranzato, Marc'Aurelio and Rohrbach, Marcus and Elhoseiny, Mohamed},
  journal={arXiv preprint arXiv:1812.00420},
  year={2018}
}

@article{chaudhry2019tiny,
  title={On tiny episodic memories in continual learning},
  author={Chaudhry, Arslan and Rohrbach, Marcus and Elhoseiny, Mohamed and Ajanthan, Thalaiyasingam and Dokania, Puneet K and Torr, Philip HS and Ranzato, Marc'Aurelio},
  journal={arXiv preprint arXiv:1902.10486},
  year={2019}
}

@article{golkar2019continual,
  title={Continual learning via neural pruning},
  author={Golkar, Siavash and Kagan, Michael and Cho, Kyunghyun},
  journal={arXiv preprint arXiv:1903.04476},
  year={2019}
}

@article{hung2019compacting,
  title={Compacting, picking and growing for unforgetting continual learning},
  author={Hung, Ching-Yi and Tu, Cheng-Hao and Wu, Cheng-En and Chen, Chien-Hung and Chan, Yi-Ming and Chen, Chu-Song},
  journal={Advances in neural information processing systems},
  volume={32},
  year={2019}
}

@inproceedings{li2019learn,
  title={Learn to grow: A continual structure learning framework for overcoming catastrophic forgetting},
  author={Li, Xilai and Zhou, Yingbo and Wu, Tianfu and Socher, Richard and Xiong, Caiming},
  booktitle={International conference on machine learning},
  pages={3925--3934},
  year={2019},
  organization={PMLR}
}

@inproceedings{wang2022foster,
  title={Foster: Feature boosting and compression for class-incremental learning},
  author={Wang, Fu-Yun and Zhou, Da-Wei and Ye, Han-Jia and Zhan, De-Chuan},
  booktitle={European conference on computer vision},
  pages={398--414},
  year={2022},
  organization={Springer}
}

@ARTICLE{10643329,
  author={Zhang, Zicheng and Wu, Haoning and Zhang, Erli and Zhai, Guangtao and Lin, Weisi},
  journal={IEEE Transactions on Pattern Analysis and Machine Intelligence}, 
  title={Q-Bench$^+$+: A Benchmark for Multi-Modal Foundation Models on Low-Level Vision From Single Images to Pairs}, 
  year={2024},
  volume={46},
  number={12},
  pages={10404-10418},
  doi={10.1109/TPAMI.2024.3445770}}

@InProceedings{Wu_2024_CVPR,
    author    = {Wu, Haoning and Zhang, Zicheng and Zhang, Erli and Chen, Chaofeng and Liao, Liang and Wang, Annan and Xu, Kaixin and Li, Chunyi and Hou, Jingwen and Zhai, Guangtao and Xue, Geng and Sun, Wenxiu and Yan, Qiong and Lin, Weisi},
    title     = {Q-Instruct: Improving Low-level Visual Abilities for Multi-modality Foundation Models},
    booktitle = {Proceedings of the IEEE/CVF Conference on Computer Vision and Pattern Recognition (CVPR)},
    month     = {June},
    year      = {2024},
    pages     = {25490-25500}
}

@article{zhang2025multimodal,
  title={A multimodal biomedical foundation model trained from fifteen million image--text pairs},
  author={Zhang, Sheng and Xu, Yanbo and Usuyama, Naoto and Xu, Hanwen and Bagga, Jaspreet and Tinn, Robert and Preston, Sam and Rao, Rajesh and Wei, Mu and Valluri, Naveen and others},
  journal={Nejm Ai},
  volume={2},
  number={1},
  pages={AIoa2400640},
  year={2025},
  publisher={Massachusetts Medical Society}
}

@inproceedings{wang2024multimodal,
  title={Multimodal llm enhanced cross-lingual cross-modal retrieval},
  author={Wang, Yabing and Wang, Le and Zhou, Qiang and Wang, Zhibin and Li, Hao and Hua, Gang and Tang, Wei},
  booktitle={Proceedings of the 32nd ACM International Conference on Multimedia},
  pages={8296--8305},
  year={2024}
}

@article{sun2025cpathagent,
  title={CPathAgent: An Agent-based Foundation Model for Interpretable High-Resolution Pathology Image Analysis Mimicking Pathologists' Diagnostic Logic},
  author={Sun, Yuxuan and Si, Yixuan and Zhu, Chenglu and Zhang, Kai and Shui, Zhongyi and Ding, Bowen and Lin, Tao and Yang, Lin},
  journal={arXiv preprint arXiv:2505.20510},
  year={2025}
}

@inproceedings{radford2021learning,
  title={Learning transferable visual models from natural language supervision},
  author={Radford, Alec and Kim, Jong Wook and Hallacy, Chris and Ramesh, Aditya and Goh, Gabriel and Agarwal, Sandhini and Sastry, Girish and Askell, Amanda and Mishkin, Pamela and Clark, Jack and others},
  booktitle={International conference on machine learning},
  pages={8748--8763},
  year={2021},
  organization={PmLR}
}

@article{achiam2023gpt,
  title={Gpt-4 technical report},
  author={Achiam, Josh and Adler, Steven and Agarwal, Sandhini and Ahmad, Lama and Akkaya, Ilge and Aleman, Florencia Leoni and Almeida, Diogo and Altenschmidt, Janko and Altman, Sam and Anadkat, Shyamal and others},
  journal={arXiv preprint arXiv:2303.08774},
  year={2023}
}

@article{abdin2024phi,
  title={Phi-4 technical report},
  author={Abdin, Marah and Aneja, Jyoti and Behl, Harkirat and Bubeck, S{\'e}bastien and Eldan, Ronen and Gunasekar, Suriya and Harrison, Michael and Hewett, Russell J and Javaheripi, Mojan and Kauffmann, Piero and others},
  journal={arXiv preprint arXiv:2412.08905},
  year={2024}
}

@inproceedings{kirillov2023segment,
  title={Segment anything},
  author={Kirillov, Alexander and Mintun, Eric and Ravi, Nikhila and Mao, Hanzi and Rolland, Chloe and Gustafson, Laura and Xiao, Tete and Whitehead, Spencer and Berg, Alexander C and Lo, Wan-Yen and others},
  booktitle={Proceedings of the IEEE/CVF international conference on computer vision},
  pages={4015--4026},
  year={2023}
}

@inproceedings{kumar2025gpt,
  title={GPT-Powered Virtual Assistants for Intelligent Cloud Service Management},
  author={Kumar, Santosh and Nutalapati, Pavan and Vemuri, Srinikhil Saisatya and Aida, RaviTeja and Salami, Zaeid Ajsan and Boob, Nandini Shirish},
  booktitle={2025 World Skills Conference on Universal Data Analytics and Sciences (WorldSUAS)},
  pages={1--6},
  year={2025},
  organization={IEEE}
}

@article{guan2024world,
  title={World models for autonomous driving: An initial survey},
  author={Guan, Yanchen and Liao, Haicheng and Li, Zhenning and Hu, Jia and Yuan, Runze and Zhang, Guohui and Xu, Chengzhong},
  journal={IEEE Transactions on Intelligent Vehicles},
  year={2024},
  publisher={IEEE}
}

@article{azad2023foundational,
  title={Foundational models in medical imaging: A comprehensive survey and future vision},
  author={Azad, Bobby and Azad, Reza and Eskandari, Sania and Bozorgpour, Afshin and Kazerouni, Amirhossein and Rekik, Islem and Merhof, Dorit},
  journal={arXiv preprint arXiv:2310.18689},
  year={2023}
}

@article{jha2024clap4clip,
  title={Clap4clip: Continual learning with probabilistic finetuning for vision-language models},
  author={Jha, Saurav and Gong, Dong and Yao, Lina},
  journal={Advances in neural information processing systems},
  volume={37},
  pages={129146--129186},
  year={2024}
}

@article{cao2024continual,
  title={Continual llava: Continual instruction tuning in large vision-language models},
  author={Cao, Meng and Liu, Yuyang and Liu, Yingfei and Wang, Tiancai and Dong, Jiahua and Ding, Henghui and Zhang, Xiangyu and Reid, Ian and Liang, Xiaodan},
  journal={arXiv preprint arXiv:2411.02564},
  year={2024}
}
\bibliographystyle{icml2026}

\clearpage
\appendix
\section{Related Works}

\textbf{Class-Incremental Learning. }The task of Class-Incremental Learning (CIL) targets to learn new classes continuously while preserving knowledge of previously learned classes~\cite{jung2020continual, hou2019learning, Kang2022afc, wang2022foster}. Several strategies have been adopted to overcome catastrophic forgetting and retain previously learned knowledge. Parameter regularization-based methods~\cite{ahn2019uncertainty, benjamin2018measuring, Chaudhry_2018_ECCV} aim to estimate the importance of different parameters of the model and allocate them with different weights to indicate their importance. Knowledge distillation-based methods~\cite{castro2018end, dhar2019learning, douillard2020podnet} help the model preserve previously learned knowledge across incremental steps by minimizing the Kullback-Leibler divergence between the output probability distributions of the previous and current models. Exemplar/Memory Replay-based methods~\cite{9710553, chaudhry2018efficient, chaudhry2019tiny, channappayya2023augmented, chen2023saving} assume that a small size of memory is accessible to store examples from old tasks/classes. Architecture-based methods~\cite{golkar2019continual, hung2019compacting, li2019learn} hold incremental modules to increase the capacity of the model to handle new tasks/classes.\\

\textbf{Foundational Multimodal Learning. }Multimodal foundation models~\cite{radford2021learning, achiam2023gpt, abdin2024phi} have demonstrated remarkable efficacy across diverse applications—ranging from low-level vision tasks~\cite{Yadav_2025_BMVC, 10643329, Wu_2024_CVPR} and image recognition~\cite{zhang2025multimodal} to cross-modal retrieval~\cite{wang2024multimodal} and agent-based reasoning~\cite{sun2025cpathagent}—by aligning and binding different modalities within a joint embedding space. Benefiting from large-scale pretraining, these architectures excel at capturing complex cross-modal semantic associations. A prominent example is the Segment Anything Model for Audio (SAM-Audio)~\cite{shi2025samaudio}, which leverages advanced latent spaces to unify audio, visual, and textual modalities for prompt-driven sound separation. However, while such models exhibit strong generalization capabilities, they face severe limitations in dynamic environments that necessitate continual adaptation. Specifically, in the challenging setting of CIL, these models remain highly susceptible to catastrophic forgetting (shown in Fig~\ref{fig:teaser}). Our objective is to harness the robust multimodal priors embedded in SAM-Audio and adapt its architecture to enable scalable, resilient class-incremental learning.
\begin{table*}[ht]
\centering
\caption{\footnotesize Comparison of our proposed approach with previous methods under different class-incremental settings on AVE-CI dataset. \textbf{Bold} denotes the best test accuracy (higher is better).}
\label{tab:results_new}
\renewcommand{\arraystretch}{1}
\resizebox{\linewidth}{!}{
\setlength{\tabcolsep}{2pt}
\begin{tabular}{lcccccccc>{\columncolor[gray]{0.9}}c}
\toprule \midrule
\multirow{2}{*}{\textbf{Settings}} & \multirow{2}{*}{\textbf{Fine-tuning}} & \textbf{LwF} & \textbf{iCaRL-NME} & \textbf{iCaRL-FC} & \textbf{SS-IL} & \textbf{AFC-NME} & \textbf{AFC-LSC} & \textbf{AV-CIL} & \textbf{Ours} \\
 &  & \textbf{~\cite{8107520}} & \textbf{~\cite{8100070}} & \textbf{~\cite{8100070}} & \textbf{~\cite{9710553}} & \textbf{~\cite{Kang2022afc}} & \textbf{~\cite{Kang2022afc}} & \textbf{~\cite{pian2023audio}} & \\
\midrule
7 classes $\times$ 4 steps & 42.40 & 58.07 & 56.15 & 65.88 & 61.94 & 68.46 & 65.21 & 74.04 & \textbf{88.72} \\
4 classes $\times$ 7 steps & 37.54 & 50.25 & 60.02 & 65.50 & 65.29 & 68.61 & 61.93 & 71.76 & \textbf{86.36} \\ \midrule
\bottomrule
\end{tabular}
}
\end{table*}

\section{Additional Results and Analysis}

\begin{figure}[h]
\centering
\includegraphics[width=\linewidth]{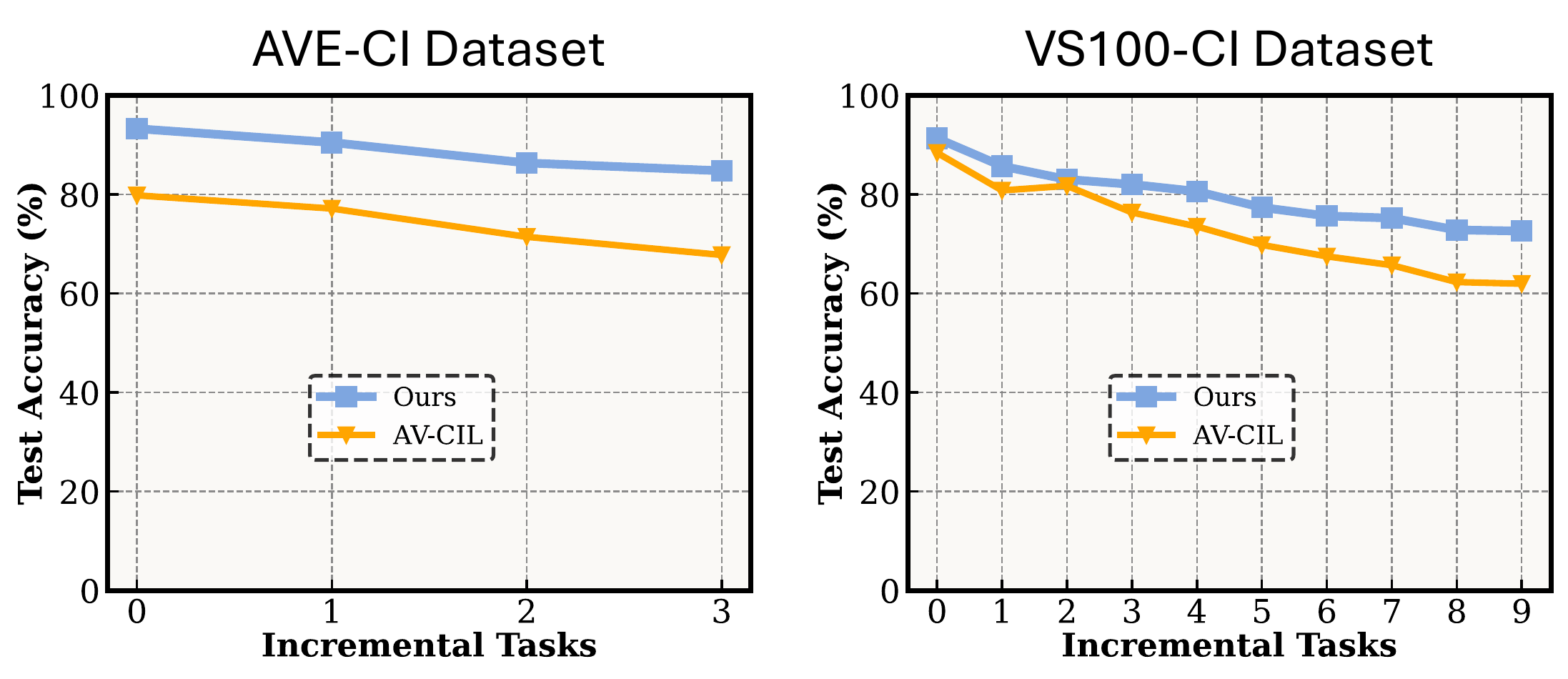}
  \centering
\caption{\footnotesize Performance (test accuracy) comparison with AV-CIL~\cite{pian2023audio}. \textbf{[Left]} Individual task performance for AVE-CI dataset on 4 incremental tasks. \textbf{[Right]} Individual task performance for VS100-CI dataset on 10 incremental tasks.}
\label{fig:analysis}
\end{figure}

In Fig.~\ref{fig:analysis}, we comparatively analyze the test accuracy of the proposed approach with AV-CIL~\cite{pian2023audio} on AVE-CI and VS100-CI at individual incremental steps. From the plots, we observe that our proposed approach consistently outperforms AV-CIL across all the incremental tasks in both benchmark datasets.

To further assess the robustness of our proposed approach, we perform an additional experiment on different class-incremental tasks. Following~\cite{pian2023audio}, we provide a new task setting with 4 distinct classes in each task, making a total of 7 tasks. This setting makes the problem more challenging with more sequential steps. As observed in Table~\ref{tab:results_new}, our proposed approach outperforms previous methods by a significant margin. Crucially, we observe a small drop in test accuracy compared to the traditional 4-step setting. We find that this drop is mainly due to an increase in the number of incremental steps.

\section{Additional Ablation Studies}
\textbf{Importance of Guided Attention. }While SAM-Audio provides a strong foundational model with highly dense representations, naively fusing its audio and visual modalities fails to adapt effectively in a class-incremental scenario. This simplistic fusion not only restricts overall accuracy but also triggers severe catastrophic forgetting as new tasks are introduced. The empirical vulnerability of naive fusion is demonstrated in Table~\ref{tab:ablation_1} [row 1], which highlights a rapid deterioration of previously acquired concepts. To circumvent this, we employ our proposed guided attention alongside auxiliary feature-level distillation losses to preserve and integrate knowledge. The inclusion of guided attention significantly bolsters performance and stabilizes the model against forgetting [row 5]. The corresponding task-wise accuracy for this ablation is further visualized in Fig.~\ref{fig:teaser}(b).
\end{document}